\documentclass[11pt]{article}

\usepackage[preprint]{acl}

\usepackage{times}
\usepackage{latexsym}

\usepackage[T1]{fontenc}

\usepackage[utf8]{inputenc}

\usepackage{microtype}

\usepackage{threeparttable}
\usepackage{inconsolata}

\usepackage{graphicx}
\usepackage{amssymb}
\usepackage{bbm}
\usepackage{booktabs}
\usepackage{amsmath}
\usepackage{tabularray}
\usepackage{enumitem}

\title{Transformer-Encoder Trees for Efficient Multilingual Machine Translation and Speech Translation}

\author{Yiwen Guan and Jacob Whitehill \\
  Worcester Polytechnic Institute \\
  \texttt{yguan2@wpi.edu, jrwhitehill@wpi.edu} \\}

\begin{document}
\maketitle
\begin{abstract}

Multilingual translation suffers from computational redundancy, especially when translating into multiple languages simultaneously. In addition, translation quality can suffer for low-resource languages. 
To address this, we introduce Transformer Encoder Tree (TET), a hierarchical, non-autoregressive encoder-only architecture trained with Connectionist Temporal Classification (CTC) for multilingual translation. 
TET shares intermediate representations among linguistically similar target languages, improving accuracy on low-resource languages while reducing computational redundancy and enabling the generation of all target languages in a single forward pass. 
TET eliminates the sequential bottleneck of autoregressive models and supports fully parallel decoding of all tokens across all target languages. 
Compared to a naive one-to-many multilingual design, TET reduces the total parameter count by 66\% and lowers inference computation by 60\%. 
In speech translation, combining TET with a non-autoregressive speech recognition backbone (Wav2Vec2) shows competitive translation quality compared to autoregressive systems while speeding up inference by approximately 7-14$\times$. 
\end{abstract}


\section{Introduction}

\begin{figure}[t]
\begin{center}
\includegraphics[width=\columnwidth]{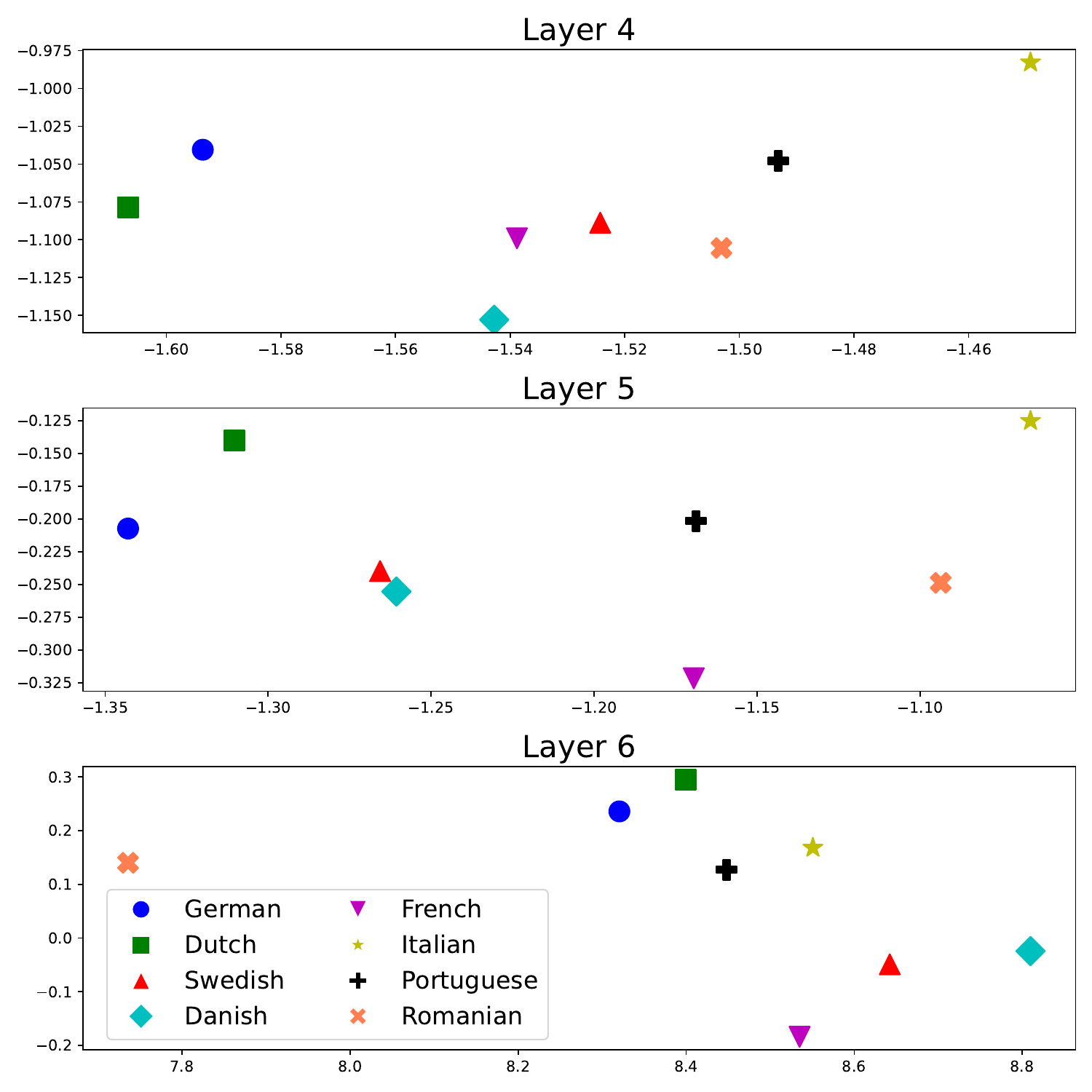}
\caption{Embedding clusters for last 3 layers in one EN-X multilingual translation model, which is a Transformer encoder-only model trained on 8 target languages, with a task token prepended to the input representing the language. The embeddings are averaged over the same 5792 sentences from Multi30K. The embeddings cluster according to their linguistic similarity. }
\label{fig:embeddings}
\end{center}
\end{figure}

We consider how to translate efficiently from a single source language into multiple target languages simultaneously and in real time. Our motivating application is what we call the \emph{United Nations task}, in which both latency and accuracy of the translation are important. Translating from the source language into each of the target languages one-at-a-time introduces substantial computational redundancy, as this approach completely ignores the relationships among target languages. For example, French, Spanish, and Italian all belong to the Romance family, and thus their latent representations learned by a multilingual model can be very close to each other (see Figure \ref{fig:embeddings}). The run-time costs of ignoring linguistic similarity become increasingly severe when the source language needs to be translated into many languages at once. 

The state-of-the-art approach to these tasks is based on autoregression, including encoder-decoder Transformers (e.g., Whisper \cite{radford2023robust}) and large language models \cite{brown2020language,achiam2023gpt,huang24h_interspeech}, offer excellent accuracy. 
However, autoregressive (AR) methods are prone to hallucination and suffer from high latency due to their inherent sequential decoding mechanism, which restricts the model to unidirectional context and limits parallel processing capabilities. 
To address these issues, recent works \cite{inaguma2021orthros, peng2024owsm, ma2024non} have explored how non-autoregressive (NAR) models can be used for efficient and accurate machine and speech translation. Building upon this line of work, we address the United Nations task by developing a hierarchical Transformer encoder-only model for multilingual machine and speech translation. By utilizing a tree structure, our proposed model capitalizes on the linguistic similarity between target languages to improve both accuracy and speed. 


In this work, we investigate how a Transformer Encoder Tree (TET) can be used for machine translation (MT), speech-to-text translation (S2TT), and speech-to-speech translation (S2ST) to efficiently perform one-to-many multilingual translation. 
Akin to ideas in \citet{azpiazu2020framework} and \citet{khusainova2021hierarchical}, we leverage language trees to build better model architectures based on linguistic similarities. 
However, these works rely on typical AR encoder-decoder architectures; our approach is encoder-only, thus enabling parallel decoding, significantly speeding up translation. 
More importantly, our approach uniquely enables the simultaneous generation of all target languages in a single forward pass. 
By constructing a hierarchical tree of encoder-only layers, the model computes shared intermediate representations of linguistic similarity among related target languages. 
These representations can be shared according to the degree of kinship, thus circumventing redundant computation. 
Furthermore, experiments demonstrate that our method can benefit low-resource languages. Our framework exhibits great potential for efficient multilingual translation applications. 

Our contributions are summarized as follows: 
\begin{itemize}
    \item We propose TET, a novel multilingual translation system based on non-autoregressive encoder-only models using Connectionist Temporal Classification (CTC) \cite{graves2006connectionist}. Our work contributes to the recent growing field of non-autoregressive approaches for translation tasks.
    \item Inspired by common feature representations, we devise an approach of structuring the translation process hierarchically. By constructing a tree model, linguistically similar target languages efficiently share intermediate representations, improving accuracy for low-resource languages and enabling parallel generation of all target languages. 
    \item By employing an encoder-only design, TET can generate all target language tokens simultaneously in one forward pass, alleviating computational redundancy across languages and eliminating the sequential bottleneck of conventional autoregressive models \cite{gu2018non}, thereby significantly reducing both latency and overall computational cost. 
\end{itemize}


\section{Background}


\subsection{Non-autoregressive Translation}
Conventional autoregressive methods generate tokens sequentially from left to right, entail high latency. 
Non-autoregressive (NAR) models generate all target tokens simultaneously, greatly accelerating inference \cite{gu2018non, libovicky2018end, kasai2021deep, xiao2023survey}. 
In contrast to  autoregressive methods, NAR models are based on a conditional independence assumption among target tokens: 
\begin{equation}
    P(Y|X;\theta)=\prod_t^TP_t(y_t|X;\theta),
\end{equation}
where $\theta$ is the model parameters, and $P_t$ is the probability in position $t$. 

Recent research has explored applying NAR methods to translation tasks to improve generation speed while preserving translation quality, such as iterative decoding, glancing sampling, or diffusion \cite{lee2018deterministic, qian2021glancing, chen2023xdlm}. 
The NAR models can significantly speed up the inference process with parallel decoding, while the vanilla NAR translation method always lags behind its autoregressive counterpart in translation quality \cite{xu2023ctc}.

\subsection{CTC-based Non-autoregressive Translation} 
Connectionist Temporal Classification (CTC) \cite{graves2006connectionist} is a sequence modeling algorithm designed to learn monotonic alignments between sequences by introducing a blank token $\epsilon$ to handle length mismatches.
Although originally designed for the chronological nature of automatic speech recognition (ASR) \cite{graves2014towards, chen2020non, lee2021intermediate, cao2025m}, recent works have successfully extended CTC to NAR translation tasks by mapping source segments to target tokens \cite{shao2022non, inaguma2021orthros, xu2023ctc}. 
CTC is a natural fit for NAR architectures because it reduces subsequences of repeated tokens into a single token. This mechanism mitigates the token repetition issues caused by the conditional independence assumption, and enables NAR models to generate sequences robustly. 

Notably, OWSM-CTC \cite{peng2024owsm} proposes a unified CTC-based encoder-only model for multilingual ASR, speech translation, and language identification, improving both accuracy and latency. Other recent advances based on CTC include \citet{ma2024non} for end-to-end simultaneous speech-to-any translation, and for textless speech-to-speech translation using discrete units \cite{fang2024ctc}. 
Despite these advancements, no prior CTC-based system can generate multiple target languages simultaneously. Although running separate NAR models for each language concurrently could achieve similar decoding speed, this approach scales linearly in both computation and memory with the number of languages, making it highly inefficient. 

Furthermore, applying CTC to translation introduces inherent challenges. Unlike ASR translation often requires the word reordering, which violates CTC's monotonic alignment and conditional independence assumptions. Various techniques mitigate these issues by exploring non-monotonic latent alignments or glancing training \cite{shao2022non, qian2021glancing}. Unlike these work aiming improving translation quality for CTC-based models, we focus on a different bottleneck: efficient simultaneous translation of multiple languages. Therefore, our work is orthogonal to CTC-specific enhancements and could theoretically be integrated with them to further boost performance. 

\subsection{Hierarchical Architectures for Multilingual Machine Translation}
 \citet{azpiazu2020framework} and \citet{khusainova2021hierarchical} exploit typological language family trees to guide parameter sharing. Their results confirm that this hierarchical topology facilitates knowledge transfer among linguistically similar languages, and is particularly beneficial for low-resource languages. However, these works employed encoder-decoder architectures, which fundamentally limit their parallelism. 



\begin{figure*}[t]
\begin{center}
\includegraphics[width=\textwidth]{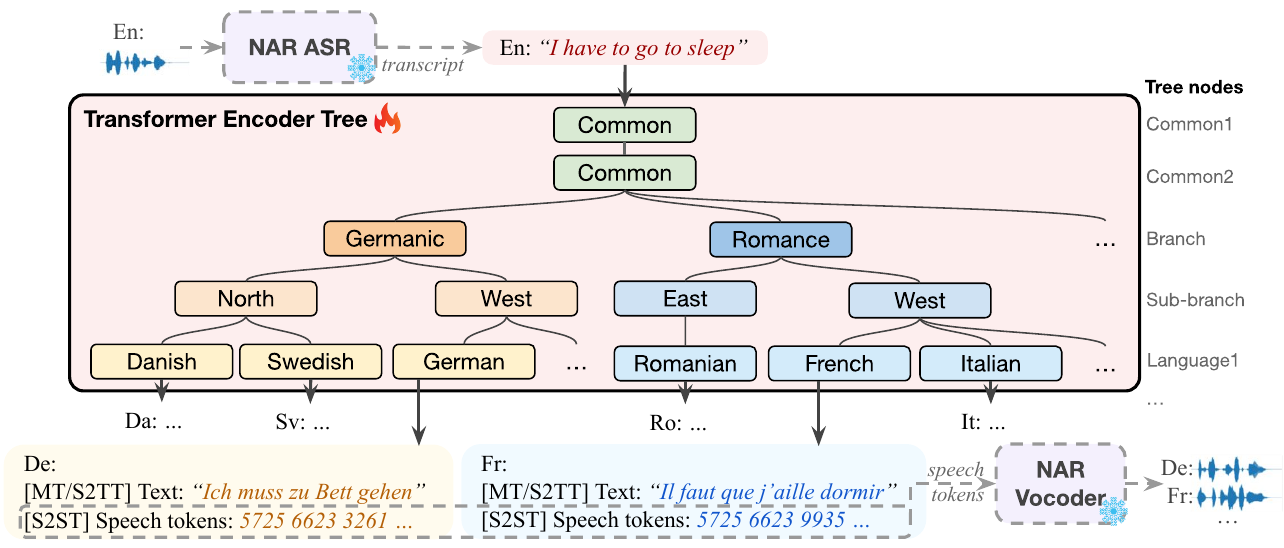}
\caption{Illustration of the MT/ST pipeline using Transformer-Encoder Tree (TET); the box in red shows the general architecture of TET for the Indo-European family. Each node in the tree represents one Transformer-encoder layer. 
An English sentence (or transcript generated by the ASR module) is passed to TET for processing; then either (1) sentences of all target languages are generated for MT task, or (2) speech tokens of all target languages are generated for S2ST task. 
The components with dashed lines are for speech translation tasks and are optional in the TET pipeline;  
we use Wav2Vec2 in NAR ASR module, and HiFi-GAN as NAR TTS vocoder. }
\label{fig:transformer_tree}
\end{center}
\end{figure*}

\section{Method}
We propose a Transformer Encoder Tree (TET) for multilingual translation that capitalizes on shared linguistic structure. Our inspiration is shown in Figure \ref{fig:embeddings}: 
 When training a single Transformer-encoder model for all EN-X translation tasks, the learned embeddings spontaneously cluster by linguistic similarity  (e.g., Germanic languages vs. Romance languages). 
Inspired by this observation, we consider incorporating linguistic knowledge at the initial stage to facilitate the training of multilingual translation models.

\subsection{Transformer Encoder Trees}
We design a hierarchical model that mimics the language tree to translate from a single source language into multiple targets, as shown in Figure \ref{fig:transformer_tree}. 

To translate from the source language into multiple target languages, we construct a TET, where each node in the tree consists of a Transformer encoder layer. 
The root and first few nodes of the tree are shared across languages (i.e., Common1 and Common2), which take source language tokens as input; the intermediate nodes are grouped into branches, sub-branches, etc., based on linguistic similarity; and each leaf node (representing a target language) outputs the corresponding target language tokens. Since TET only contains Transformer encoders, it is a non-autoregressive method. Hence, by reusing the activations of the nodes in the tree, we can significantly reduce the total computation needed to translate the source language into all target languages.

To give an example, consider translating an English input sentence into 8 target languages in \{De, Nl, Sv, Da, Fr, It, Pt, Ro\} (refer to language codes in Table \ref{tab:tatoeba_size}). 
To begin with, they can be divided into Germanic languages: \{De, Nl, Sv, Da\} and Romance languages: \{Fr, It, Pt, Ro\}. Next, they can be further divided into sub-branches like Northern Germanic: \{Da, Sv\}, Western Germanic: \{De, Nl\}, etc. 
In this way, only $2 \text{ (\textit{common})}+ 2 \text{ (\textit{branch})}+ 4 \text{ (\textit{sub-branch})} + 8 \text{ (\textit{language})}=16$ layers need to be processed in TET, instead of $8 \text{ (\textit{language})}\times 5 \text{ (\textit{height})}=40$ layers for a naive approach where each language receives its own sequence of layers.
In this example, $(40-16)/40=60\%$ of the computation is saved thanks to the tree structure design. 

Similar to prior work on encoder-only speech foundation model \cite{peng2024owsm}, we train TET by optimizing the standard CTC loss defined by: 
\begin{equation}
    \mathcal{L}_\text{CTC}=-\text{log}\ P_\text{CTC}(Y|X),
\end{equation}
where $P_\text{CTC}(Y|X)$ is the conditional probability of the target sequence $Y$ marginalized over all valid alignments given the input $X$. 
During inference, we use greedy search to find the alignment with the highest probability. 
To ensure that the model has sufficient space for ``blank'' tokens, random padding is added to each input sequence, and the model's predictions are post-processed with CTC collapse. The input and output format is exemplified in Table \ref{table:format}.

\begin{table}[t]
\centering
\setlength{\tabcolsep}{2pt}
\resizebox{0.9\columnwidth}{!}{%
\begin{tabular}{l|l}
        \toprule
        Inputs (raw)         & "GOOD DAY" \\
                             & "HOW ARE YOU" \\ 
        \midrule
        Inputs (random padded)  & "-G-OO- -D- D-AY- -" \\
                             & "H- -OW -A-RE Y-OU" \\ 
        \midrule
        Predictions (raw)   & "BB-O-NN- - -JO-UUR" \\
                             & "C-OM-E-T ÇA VVA-" \\ 
        \midrule
        Predictions (CTC-mapped) & "BONJOUR"\\
                             & "\textcolor{red}{COMET} ÇA VA" \\
        \midrule\midrule
        \textbf{Ground Truth}    & "BONJOUR" \\
                             & "COMMENT ÇA VA" \\ 
        \bottomrule
\end{tabular}
}
\caption{Input and output format of TET. Here, ``-'' denotes the CTC blank token $\epsilon$.}
\label{table:format}
\end{table} 

\subsection{Speech Processing Components}

Beyond machine translation, we explore how TET can be used for speech-to-text translation (S2TT) and speech-to-speech translation (S2ST). 

We adopt a cascaded approach that first transcribes source-language audio with an ASR component, then feeds the transcript into the TET to generate target-language sentences. 
Specifically, to examine the benefit of encoder-only architectures, we employ off-the-shelf ASR models with different intrinsic architectures (i.e., AR vs. NAR). 

As for S2ST, TET predicts discrete speech units rather than text tokens. We convert the discrete speech units produced by TET back to waveform using the SeamlessM4T HiFi-GAN–based text-to-speech (TTS) vocoder \cite{barrault2023seamlessm4t}. See Figure \ref{fig:transformer_tree} for more details.


\section{Experimental Setup}
To evaluate the effectiveness of the proposed model, we compare it against several competing methods for both machine translation and speech translation. 

\subsection{Baselines and Variants}
For machine translation, we compare TET to the following architectures (see Table \ref{tab:baselines}):
\begin{itemize}
    \item {\bf TET-Rnd}: A tree with the same topology as TET, but the languages are randomly assigned to leaf nodes rather than following the language tree topology;
    \item  {\bf TEnc/lang}: One Transformer encoder per language, with the same depth as TET;
    \item  {\bf TEnc/all}: A single Transformer encoder for all languages, with the same depth as TET, using a special token to specify the target language. 
\end{itemize}

For speech translation, we compare to the non-autoregressive {\bf Wav2Vec2} \cite{baevski2020wav2vec} as well as the autoregressive {\bf Whisper} \cite{radford2023robust} models as the ASR component in the TET pipeline (see the upper-left part in Figure \ref{fig:transformer_tree}): they produce English transcriptions of the input audio, then the English sentences are fed into TET for one-to-many translation.

\begin{table}[t]
\centering
\setlength{\tabcolsep}{3pt}
\resizebox{\columnwidth}{!}{%
    \begin{tabular}{lccc}
        \toprule
        \textbf{Model} & Structured sharing& Linguistic prior& Lang-specific capacity\\
        \midrule
        TET& $\checkmark$& $\checkmark$& $\checkmark$\\
         TET-Rnd& $\checkmark$& $\times$&$\checkmark$\\
        TEnc/lang& $\times$& $\times$& $\checkmark$\\
         TEnc/all& $\times$& $\times$&$\times$\\
        \bottomrule
    \end{tabular}
}
\caption{Comparison of structural properties across the proposed TET and baseline models.} 
\label{tab:baselines}
\end{table}

\subsection{Datasets}

We evaluate the proposed TET model on two datasets: Multi30K \cite{elliott2016multi30k} and Tatoeba \cite{tiedemann2020tatoeba}, as described below.  

\paragraph{Multi30K} contains image captions in English as well as their German translations. To obtain the parallel sentences of other languages, we use OpenAI's GPT-4-nano with the prompt, \textit{``Translate the sentence below into all 8 different languages (Danish, Dutch, French, German, Italian, Portuguese, Romanian, and Swedish). Precede each translation with the language name. Sentence: \ldots''}. Example sentences include: (1) ``Several men in hard hats are operating a giant pulley system.'', (2) ``A little girl climbing into a wooden playhouse.'', and (3) ``A man in a blue shirt is standing on a ladder cleaning a window.''
In our extended Multi30k dataset, each sentence is available in all 9 languages (En, De, Da, Nl, Fr, It, Pt, Ro, and Sv). 
Note that although Multi30K is a multimodal dataset, we only conduct text-only translation so no image is involved.

\begin{table}[t]
\centering
\setlength{\tabcolsep}{2pt}
\resizebox{\columnwidth}{!}{%
    \begin{tabular}{lr|lr|lr}
        \toprule
        Danish (Da) & 24,087 & Dutch (Nl) & 59,729 & French (Fr) & 176,725 \\
        \midrule
        German (De) & 223,435 & Italian (It) & 199,450 & Portuguese (Pt) & 155,821 \\ 
        \midrule
        Romanian (Ro) & 12,728 & Polish (Pl) & 45,842 & Russian (Ru) & 341,335 \\
        \midrule
        Spanish (Es) & 163,930 & Swedish (Sv) & 20,398 & & \\
        \bottomrule
    \end{tabular}
}
\caption{Number of sentence pairs between English and each target language on processed Tatoeba (En-X).}
\label{tab:tatoeba_size}
\end{table}

\paragraph{Tatoeba} is a large multilingual dataset for language learning, linguistics, and machine translation research. Since some Tatoeba examples actually contain multiple concatenated sentences or embedded quotations, we filtered it to remove examples containing any of the following characters: \begin{tt},.?!"\end{tt}. 
In the end, we harvest 637,833 unique English sentences from Tatoeba that contain translations to any of the 11 languages listed in Table \ref{tab:tatoeba_size}. 
Example sentences include: (1) ``It almost scared me not to see you online for a whole day.'', (2) ``You must be worn out after working all day.'', and (3) ``It was proven that fleas living on dogs jump higher than fleas living on cats.''
Specifically, for speech translation, we select a subset of examples that are available in all 5 high-resource target languages (Fr, De, It, Ru, Es), and form a subset of 22k examples.
Since the original Tatoeba dataset does not contain speech, we use an open-source text-to-speech converter, pyttsx3 \footnote{\url{https://github.com/nateshmbhat/pyttsx3}}, to synthesize speech utterances of each language, and convert them into discrete speech units with SeamlessM4T \cite{barrault2023seamlessm4t}.


\begin{table*}[ht]
\centering
\setlength{\tabcolsep}{4pt}
\resizebox{0.9\textwidth}{!}{%
    \begin{tabular}{l|cccccccc|cccccccc|cc}
    \toprule
         & \multicolumn{16}{c}{\bf Target Language} & \multicolumn{2}{c}{ } \\
     \cmidrule{2-19}
         & \multicolumn{8}{c|}{\bf BLEU} & \multicolumn{8}{c|}{\bf COMET} & \multicolumn{2}{c}{\bf Avg. } \\
    {\bf Model} & De& Nl& Sv& Da& Fr& It& Pt& Ro& De& Nl& Sv& Da& Fr& It& Pt& Ro& BLEU & COMET \\ 
    \midrule
    \textbf{TET} (Ours) & 40.6 & 41.1 & 49.9 & 51.0 & 44.2 & 37.4 & 39.2& 32.6 & 0.66& 0.68& 0.78& 0.77& 0.69& 0.68& 0.70& 0.68& 42.0 & 0.70\\ 
    TET-Rnd & 39.0 & 38.6 & 49.3 & 50.0 & 40.7 & 35.8 & 38.7 & 30.6& 0.65& 0.67& 0.76& 0.76& 0.68& 0.66& 0.69& 0.67& 40.3 & 0.69\\ 
    TEnc/lang & 40.0 & 41.6 & 50.2 & 49.9 & 44.1 & 40.1 & 46.2 & 36.7 & 0.66 & 0.70 & 0.80 & 0.78 & 0.70& 0.70& 0.74& 0.69& 43.6 & 0.72\\ 
    TEnc/all & 39.1 & 40.9 & 49.1 & 48.3 & 44.0 & 39.0 & 42.4 & 36.4 & 0.66& 0.70& 0.78& 0.77& 0.69& 0.70& 0.72& 0.72& 42.4 & 0.72\\ 

    \bottomrule
    \end{tabular}
}
\caption{EN-X translation performance on Multi30K (SacreBLEU$\uparrow$ / COMET$\uparrow$).}
\label{table:multi30k}
\end{table*}


\begin{table*}[ht]
\centering
\setlength{\tabcolsep}{2pt}
\resizebox{\textwidth}{!}{%
    \begin{tabular}{l|ccccccccccc|ccccccccccc|cc}
    \toprule
            & \multicolumn{22}{c}{\bf Target Language}  & \multicolumn{2}{c}{ } \\ 
     \cmidrule{2-25}
         & \multicolumn{11}{c|}{\bf BLEU} & \multicolumn{11}{c|}{\bf COMET} & \multicolumn{2}{c}{\bf Avg. } \\
    {\bf Model} & Da & Nl & Fr & De & It & Pt& Ro& Pl& Ru & Es & Sv & Da & Nl & Fr & De & It & Pt& Ro& Pl& Ru & Es & Sv & BLEU & COMET \\
    \midrule
    \textbf{TET} (Ours) & 29.5 & 20.1 & 22.8 & 13.1 & 36.1 & 29.2 & 10.2 & 9.8 & 23.1 & 24.2 & 22.7 
& 0.72& 0.67& 0.65& 0.58& 0.76& 0.74& 0.59& 0.65& 0.71& 0.69& 0.69& 21.9 & 0.68\\ 
    TET-Rnd & 26.1 & 18.8 & 19.7 & 13.2 & 32.9 & 26.7& 8.6 & 10.3 & 23.1 & 22.6 & 23.4 
& 0.71& 0.66& 0.63& 0.57& 0.75& 0.72& 0.58& 0.64& 0.70& 0.68& 0.70& 20.5 & 0.67\\ 
    TEnc/lang & 30.2 & 23.5 & 28.2 & 14.0 & 24.3 & 36.2 & 1.6 & 12.7 & 30.8 & 28.8 & 19.2 
& 0.73& 0.71& 0.69& 0.60& 0.72& 0.80& 0.46& 0.66& 0.77& 0.73& 0.68& 22.7 & 0.69\\ 
    TEnc/all & 27.7 & 17.1 & 16.9 & 10.0 & 26.6 & 23.7 & 11.2 & 9.8 & 16.3 & 20.9 & 22.0 & 0.72& 0.66& 0.62& 0.55& 0.73& 0.72& 0.62& 0.63& 0.66& 0.68& 0.69& 18.4 & 0.66\\
    \bottomrule
    \end{tabular}
}
\caption{EN-X translation performance on Tatoeba (SacreBLEU$\uparrow$ / COMET$\uparrow$).}
\label{table:tatoeba_mt}
\end{table*}

\begin{table*}[ht]
\centering
\setlength{\tabcolsep}{3pt}
\resizebox{0.85\textwidth}{!}{%
\begin{threeparttable}
    \begin{tabular}{l|c|c||ccccc|ccccc|cc}
    \multicolumn{15}{c}{\bf {\bf S2T Translation Accuracy (SacreBLEU$\uparrow$ / COMET$\uparrow$): Tatoeba (subset of 4392 utterances)}} \\ 
    \toprule
     \multicolumn{3}{c||}{} &\multicolumn{10}{c}{\bf Target Language}& \multicolumn{2}{c}{ } \\
     \cmidrule{4-15}
     \multicolumn{3}{c||}{ }    & \multicolumn{5}{c|}{\bf BLEU} & \multicolumn{5}{c|}{\bf COMET} & \multicolumn{2}{c}{\bf Avg. } \\
    {\bf ASR Model}& {\bf Size} & {\bf RTFx} & Fr & De & It & Ru & Es & Fr & De & It & Ru & Es & BLEU & COMET \\ 
    \midrule
    None (MT) \tnote{$\dagger$} &  -- & --  & 30.7 & 20.9& 35.1 & 22.9& 32.9& 0.73 & 0.67& 0.79 & 0.77& 0.78 & 28.5 & 0.75\\ 
    \midrule\midrule
    \multicolumn{9}{l}{\textbf{Whisper}  (encoder-decoder) \cite{radford2023robust}} \\
    small.en & 244M & 24.8& 29.1& 20.1& 32.9& 20.8& 31.0 & 0.72 & 0.66 & 0.78 & 0.75 & 0.76& 26.8 & 0.74\\ 
    medium.en & 769M & 13.4& 29.5& 20.2& 33.1& 20.7& 30.9 & 0.72 & 0.66 & 0.78 & 0.76 & 0.77& 26.9 & 0.74\\ 
    large-v2 & 1.55B & 8.7& 29.1& 19.8& 32.7& 20.8& 30.8 & 0.72 & 0.66 & 0.78 & 0.76 & 0.77& 26.7 & 0.74\\  
    \midrule\midrule
    \multicolumn{9}{l}{\textbf{Wav2Vec2} (encoder-only) \cite{baevski2020wav2vec}} \\
    base-960h & 95M & 196.0& 27.6 & 19.4& 30.5& 20.3& 29.1 & 0.71 & 0.65 & 0.76 & 0.74 & 0.75& 25.4 & 0.72\\ 
    large-960h & 317M & 158.9& 28.3 & 19.6& 31.7& 20.4& 30.1 & 0.72 & 0.66 & 0.77 & 0.75 & 0.76& 26.0 & 0.73\\ 
    \bottomrule
    \end{tabular}
    
    \begin{tablenotes}
      \small
      \item[$\dagger$] The theoretical upper bound where the ground truth ASR transcript is used as model input for machine translation.
    \end{tablenotes}
\end{threeparttable}
}

\caption{Evaluation of EN-X S2T translation. Wav2vec2 and Whisper models are compared based on the throughput (RTFx$\uparrow$) and translation quality (SacreBLEU and COMET). }
\label{table:tatoeba_s2tt}
\end{table*}

\paragraph{Post-processing.} To reduce the number of unique tokens in our dataset, we further process the Multi30K and Tatoeba data by removing non-European tokens (in case the automatic translation through GPT-4-nano erroneously produces any such extraneous tokens) and also by converting all sentences into capital letters.

\subsection{Training Details}
\label{subsec:training_details}

\paragraph{Model architecture.} Each node in TET is a Transformer encoder layer having a model dimension of 768, 6 attention heads, and a feed-forward dimension of 2048. 

We use different tree settings based on datasets. 
For Multi30K, the TET model has 8 leaf nodes, each corresponding to one of the 8 target languages. Each leaf node has a depth of 5, representing a path from the root node: \textit{common$_1$--common$_2$--branch--sub-branch--language}.
Similarly, for Tatoeba, the TET model has 11 leaf nodes, each has a depth of 6, representing a path of \textit{common$_1$--common$_2$--branch--sub-branch--language$_1$--language$_2$}. 
The total model size of TET is 97M on Multi30K, and 191M on Tatoeba. 

In each minibatch, we iterate through all target languages, obtain the corresponding translations and model predictions, and finally update the relevant model weights. 
To clarify with an example, suppose that there are two languages, and the TET consists of one shared encoder layer $S$ with two target languages and two language-specific leaf nodes $L_1$ and $L_2$. Then for a given input minibatch, the model weights of layers $S$ and $L_1$ would be updated, and then the model weights of layers $S$ and $L_2$ would be updated sequentially. 
Because the CTC loss requires that prediction length to be no shorter than the target length, we pad each input sequence to be 50 tokens longer than the longest target sentence in the minibatch.
To improve generalization of model weights and positional encodings, these padding tokens are inserted randomly in the input sequence rather than only at the end (see Table \ref{table:format}). 

\paragraph{Implementation details.}
All models are trained and evaluated on one NVIDIA A100 GPU (40GB) with a batch size of 32. We use the Adam optimizer \cite{DBLP:journals/corr/KingmaB14} with $\beta=(0.9, 0.999)$ and $\epsilon=1e^{-8}$, and a learning rate of $1e^{-4}$ and $2e^{-5}$ for Multi30K and Tatoeba, respectively. All models are trained for 100 epochs on Multi30K, and 30 epochs on Tatoeba. On both datasets, we use 80\% data for training and the remaining 20\% for testing.

\subsection{Evaluation Metrics}
For all experiments, we report SacreBLEU ($\uparrow$) \cite{post-2018-call} between ground-truth sentences and predictions, and COMET ($\uparrow$) \cite{rei2020comet} based on source language sentences, references, and predictions. 
To evaluate the throughput of speech-to-text translation models, we calculate the inverse real-time factor (RTFx$\uparrow$): 
\begin{equation}
    \text{RTFx}=\frac{T_\text{audio}}{T_\text{compute}},
\end{equation}
which is defined as the ratio of audio duration to processing time, indicating how many times faster than real time the model can process audio. Larger RTFx means faster inference speed.

\section{Main Results}

\subsection{Machine Translation}
Table \ref{table:multi30k} and Table \ref{table:tatoeba_mt} compare En-X machine translation results across model architectures on Multi30K and Tatoeba.

On Multi30K, compared to the one-model-per-language (TEnc/lang) and one-model-for-all (TEnc/all), TET enables translating into 8 languages in parallel while the overall translation quality is only reduced by 1.6 (-3.7\% relative reduction) and 0.4 (-0.9\%) in terms of BLEU. 

On Tatoeba, similarly, the translation quality of TET is slightly reduced by 0.8 (-3.5\%) compared to TEnc/lang. When compared to TEnc/all, however, TET outperforms it by 3.5 (+19\%) in terms of BLEU while still being computationally efficient. 

\paragraph{Effects of shared representations.}
In addition to the aforementioned benefits of saving intermediate computations, our experiments also corroborate prior work  \cite{khusainova2021hierarchical} showing that shared representations help the translation quality of low-resource languages. 
In Table \ref{table:tatoeba_mt}, on the 2 languages with the most limited resources (i.e., Romanian and Swedish), both the TEnc/all and TET surpass the TEnc/lang models, especially with a significant improvement in the BLEU score for Romanian. While both TET and Tenc/all show reduced accuracy for high-resource languages, namely Russian and Spanish, we note that TET's accuracy on these languages suffers much less.

In addition, when comparing the two shared representation models, TET exhibits an advantage over TEnc/all in almost all languages, only lagging slightly behind in Romanian. 
This indicates that TET benefits from linguistic priors to construct a more effective knowledge-sharing topology, thereby fostering overall translation quality over the vanilla sharing strategy. 

Regarding efficiency (Table \ref{tab:efficiency}), TET enables parameter sharing across languages hence reducing computation. Compared to the naive design (TET/all), TET only requires 10.7 GFLOPs in total, yielding a 60\% reduction in computation.


\subsection{Speech-to-Text Translation}
Table \ref{table:tatoeba_s2tt} presents En-X cascaded speech-to-text translation results, comparing the Wav2Vec2 (NAR) and Whisper (AR) models for the ASR component. 
In addition to translation quality metrics, we report the inverse real-time factor (RTFx) to evaluate the speed of NAR models. 

Notably, the TET pipeline incorporating Wav2Vec2-large960h, comparable in size to Whisper-small and Whisper-medium, yields only 3\% lower average BLEU while being 7-14$\times$ faster in RTFx. 
Building on this, our fully non-autoregressive TET pipeline achieves comparable speech-to-text translation accuracy while being substantially faster through parallel decoding. When translating a source speech into 5 languages simultaneously, the entire TET pipeline can be 35-70$\times$ faster than an AR approach with a comparable parameter size. 
Although the AR pipeline can be replicated multiple times in GPU to achieve parallel multilingual translation, its RTFx still lags behind the NAR pipeline and incurs significant memory overhead. 

These results highlight TET with a non-autoregressive ASR backbone as an efficient solution for low-latency simultaneous multilingual translation. 
It is conceivable that based on our TET pipeline, an end-to-end non-autoregressive speech processing pipeline can be implemented with minimal effort.

\begin{table}[t]
\centering
\setlength{\tabcolsep}{3pt}
\resizebox{0.85\columnwidth}{!}{%
    \begin{tabular}{l|c|c}
        \toprule
        \textbf{Model} & \textbf{Total Params} (M) & \textbf{Total GFLOPs}  \\
        \midrule
        TEnc/all & 284.8 & 26.8  \\
        \midrule
        TET & 96.9 (-66\%)  & 10.7 (-60\%) \\ 
        \bottomrule
    \end{tabular}
}
\caption{Model efficiency comparison on Multi30K. We report the total parameters and floating-point operations (FLOPs) when translating into 8 languages simultaneously. TEnc/lang would have same performance as TEnc/all since the architectures are the same. } 
\label{tab:efficiency}
\end{table}

\subsection{Speech-to-Speech Translation}
We also validate our framework on a 4-language S2ST setting (De, Fr, It, and Es). For evaluation, the HiFi-GAN generated speech is transcribed with Whisper-large-v2 to compute ASR-WER and ASR-BLEU with respect to the references. The TET model gets an average WER of 79.8\% and an average BLEU score of 20.3 across all languages. We speculate that the limited performance is due to the regular CTC setup, which lacks target-side context and produces shorter outputs. Further exploration of decoding strategies and alternative loss functions is warranted.


\section{Effectiveness of Tree Topology}
Admittedly, the patterns learned by neural networks can differ from human linguistic taxonomies. 
To assess the effectiveness of the language tree structure, we compare TET with the following 2 conditions: 
\begin{itemize}[itemsep=1pt]
    \item [(1)] The TET-Rnd variant, where languages are randomly assigned to leaf nodes in TET instead of following the language family tree. 
    \item [(2)] 100 distinct variants whose tree topologies are randomly constructed, but roughly maintain a balanced tree shape. Due to the resource constraints, we train each model for 10 epochs on Multi30K. 
\end{itemize}

To better clarify condition (1), it aims to answer questions like: \textit{given the same tree structure, should German share more parameters with Dutch or Italian?} 
In Table \ref{table:multi30k} and Table \ref{table:tatoeba_mt} (TET vs. TET-Rnd), the linguistic TET consistently outperforms TET-Rnd on both datasets across almost all target languages by 4\%-7\% relatively in terms of BLEU, and 1.5\% in terms of COMET. 
This ablation experiment obviates the influence of the hierarchical structure itself, thus revealing the net benefit of prior linguistic similarity knowledge. 

\begin{figure}[t]
\begin{center}
\includegraphics[width=\columnwidth]{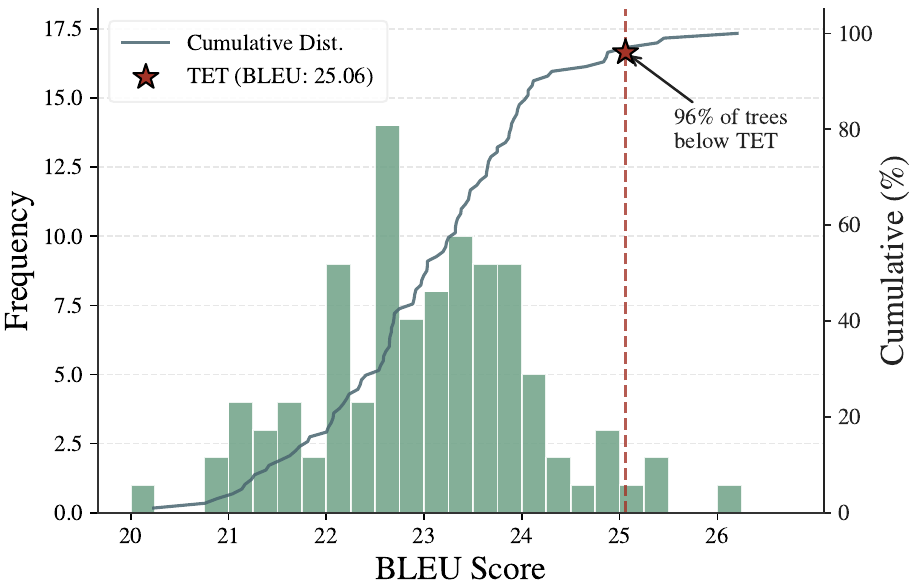}
\caption{BLEU distribution and cumulative percentage of 101 tree topologies trained for 10 epochs on Multi30K. 
The histogram shows the frequency of BLEU scores (left axis), while the solid line indicates the cumulative distribution (right axis). 
TET achieves a BLEU of 25.06, outperforming 96\% of the sampled trees (ranking 4th overall).}
\label{fig:tree_bleu}
\end{center}
\end{figure}

As for (2), it aims to answer the question: \textit{does the language tree structure really provide a better topology?} 
The statistics of the BLEU distribution of all tree topologies are shown in Figure \ref{fig:tree_bleu}. 
Among all 101 distinct tree topologies, TET ranks fourth, with a small BLEU decrease of 1.14 (-0.7\% relatively) from the top-ranked model. 
To investigate whether this performance gap persists at convergence, we extended the topology study by training the top 3 candidates alongside TET to 100 epochs. The resulting BLEU scores are 42.18, 42.01, 41.68, and 41.98 (TET), narrowing the gap to the best topology to $< 0.2$. This indicates that TET's structure-sharing design is near-optimal without requiring computationally expensive neural architecture search.

Furthermore, the top-performing structures consistently show meaningful linguistic patterns. For example, the top-ranked trees spontaneously group Romance languages (e.g., It, Pt, Ro) and Germanic languages (e.g., Da, Nl, Sv) together. These emergent clusters match the linguistic tree used in TET, indicating that linguistic similarity can effectively predict the preferred architecture of a trained neural network. 

On both datasets, the tree topology appears to have a significant impact on accuracy, highlighting the benefit of incorporating linguistic knowledge in architecture design. 
Some examples of the translation process along the TET tree path are listed in Appendix \ref{appendix:examples}.

\section{Conclusions}

In this work, we propose Transformer Encoder Tree (TET), a CTC-based NAR multilingual translation system. 
By leveraging a hierarchy based on linguistic similarity, TET efficiently shares intermediate representations, improves accuracy for low-resource languages, and improves computational efficiency of the multilingual translation process. 
Specifically, on Multi30K, TET reduces the total parameter count from 288M to 97M (-66\%) and lowers the total inference computation from 26.8 GFLOPs to 10.7 GFLOPs (-60\%), compared to a naive one-to-many multilingual design. This makes TET suitable for efficient deployment on edge devices. 
Additionally, non-autoregressive speech models (e.g., Wav2Vec2) can be easily plugged in and trained end-to-end along with TET to enable fast multilingual speech translation. 

TET is primarily designed as an efficient architecture for parameter sharing across multiple target languages, rather than another standalone NAR model. In this work, we intentionally isolate the effect of structured parameter sharing. 
Since our work is orthogonal to other NAR or CTC-based methods, general strategies for improving NAR or CTC-based translation quality can be naturally incorporated on top of TET in future work, such as non-monotonic latent alignments \cite{shao2022non}, target length prediction \cite{ghazvininejad2019mask}, reordering augmentation \cite{xu2023ctc}, and glancing sampling \cite{qian2021glancing}. 

Overall, our approach shows effectiveness and strong potential for scenarios that require accurate, low-latency machine translation and speech translation across multiple target languages. 

{\bf Future work.} 
In the future, the proposed method can be evaluated on larger machine translation and speech translation datasets, and extended to larger backbones and more diverse languages to validate scalability. 


\section*{Limitations}
While the Transformer Encoder Tree (TET) demonstrates significant improvements in latency and resource efficiency for multilingual translation, this study has certain limitations. 

Our evaluation primarily focuses on a specific set of language families and relatively controlled datasets, which may not fully represent the linguistic diversity in open-domain corpora. 

Moreover, our speech translation experiments are based on synthesized speech, which may contain inherent bias and not fully capture real-world acoustic variability. More difficult real-world datasets should be used to validate the performance of the real speech translation task. (e.g., Europarl, CoVoST2, etc.). 

While being much more efficient in inference speed, strong autoregressive models such as T5 could outperform our approach in translation quality in such settings.

From our results, TET ranks 4th among all 101 tree topologies and the performance gap from the top-ranked model is almost negligible. However, an optimal tree topology may be discovered in the future that can stably achieve the best overall accuracy. The selection of tree topologies leaves room for further optimization through automated architecture search.

\section*{Acknowledgments}

This research was supported by the  National Science Foundation (NSF) National AI Institute for Student-AI Teaming (iSAT) under grants DRL \#2454151, and also from an NSF CAREER grant \#2046505. The opinions expressed are those of the authors and do not represent views of the NSF.


\bibliography{paper}

\appendix

\section{Appendix}
\label{sec:appendix}

\subsection{Language Paths in TET}
The detailed paths of all languages in TET are shown in Table \ref{table:path}. 
\begin{table}[ht]
\centering
\setlength{\tabcolsep}{3pt}
\resizebox{\columnwidth}{!}{%
    \begin{tabular}{c|c|l}
    \toprule
    \textbf{Dataset} & \textbf{Language} & \multicolumn{1}{c}{\textbf{Path}} \\ 
    \midrule  
     & German & \textit{--Germanic--Western--German}  \\ 
     & Dutch & \textit{--Germanic--Western--Dutch}  \\ 
     & Swedish & \textit{--Germanic--Northern--Swedish}  \\ 
    Multi30K & Danish & \textit{--Germanic--Northern--Danish}  \\ 
     & French & \textit{--Romance--Western--French}  \\ 
     & Italian & \textit{--Romance--Western--Italian}  \\ 
     & Portuguese & \textit{--Romance--Western--Portuguese}  \\ 
     & Romanian & \textit{--Romance--Eastern--Romanian}  \\ 
    \midrule \midrule 
     & Danish & \textit{--Germanic--Northern--Danish$_1$--Danish$_2$}  \\ 
     & Dutch & \textit{--Germanic--Western--Dutch$_1$--Dutch$_2$}  \\ 
     & French & \textit{--Romance--Western--French$_1$--French$_2$}  \\ 
     & German & \textit{--Germanic--Western--German$_1$--German$_2$}  \\ 
     & Italian & \textit{--Romance--Western--Italian$_1$--Italian$_2$}  \\ 
    Tatoeba & Portuguese & \textit{--Romance--Western--Portuguese$_1$--Portuguese$_2$}  \\ 
     & Romanian & \textit{--Romance--Eastern--Romanian$_1$--Romanian$_2$}  \\ 
     & Polish & \textit{--Slavic--Western--Polish$_1$--Polish$_2$}  \\ 
     & Russian & \textit{--Slavic--Eastern--Russian$_1$--Russian$_2$}  \\ 
     & Spanish & \textit{--Romance--Western--Spanish$_1$--Spanish$_2$}  \\ 
     & Swedish & \textit{--Germanic--Northern--Swedish$_1$--Swedish$_2$}  \\ 
    \bottomrule
    \end{tabular}
}
\caption{Paths of all languages in TET on Multi30K and Tatoeba. All paths begin with "\textit{common$_1$--common$_2$--$\cdots$}'', which is omitted in this table for simplicity. }
\label{table:path}
\end{table}

\subsection{Translation Examples}
\label{appendix:examples}
Examples for the inference process of TET model is shown in Table \ref{tab:example0}, \ref{tab:example1} and \ref{tab:example2}. For the first example, the TET follows the Multi30K model architecture mentioned in Sec. \ref{subsec:training_details}. For the last 2 examples, the TET model is trained on Europarl \cite{koehn2005europarl} dataset from scratch with 6 Transformer encoder layers. 
The model is trained on the following target languages (from English): Dutch, Danish, French, Spanish, and Romanian. 

To display the intermediate translation results, we apply the output linear layer of each language's leaf node to all nodes on its tree path. Note that this operation is for display purposes only; during the actual inference phase, the intermediate layer nodes remain shared across languages, so the representations should be the same.  

The path of each language pair is as follows: En-Nl: \textit{common$_1$--common$_2$--Germanic--Western--Dutch$_1$--Dutch$_2$}; En-Fr: \textit{common$_1$--common$_2$--Romance--Western--French$_1$--French$_2$}; En-Es: \textit{common$_1$--common$_2$--Romance--Western--Spanish$_1$--Spanish$_2$}. 

\subsection{Datasets}
The Multi30K and Tatoeba datasets (and also Europarl in this Appendix) are publicly available for research purposes. We confirm that our use of the Multi30K and Tatoeba (and also Europarl) is consistent with their intended use for machine translation research.

\begin{table*}[ht]
\centering
\begin{tblr}{
  colspec = {Q[m, l] X[l] X[l]}, 
  width = \textwidth,
  hlines = {0.3pt},
  hline{1,Z} = {1.2pt},
  vlines = {0.3pt},
  vline{1,Z} = {0pt},
  rows = {t},
  row{2-Z} = {font = \small},
  column{1} = {font=\normalsize},
}
    \textbf{Tree Node} & \textbf{German} & \textbf{French}  \\
    Common1 & "ÃXẠQOŌSYSPLÛ ... ÈTGLÛLCETQC PHLCZT" & "ŒWXJВŒÎVÅXÂŒÂ ... À CLÔXYCÔÛIËHfF"  \\
    Common2 & "EX QOÑÛLÉJÈZ JÄIYT PÜYÄET DIM JÛBŌXG JÈFHT JWÄCT DẠÑ QÜAILZDY" & "ŒW ВНКÚËНÀJQKEПJTJYÇANZÙ LAÍ JÔÛZWÀ JQÔÈQKEÀ JÔÇNEÙ LAÍXÀ BÚÔÏLKÉX" \\
    Branch & "EX QOSTBALPJÉ ÜY YSPÜYÄT DI GÛBUS JÈÄAÉ ÜWÜT DIÑ ZKLALY" & "ŒÙEW JQUQZÚDÀ JÚÔBKE JETJ JЕ ÀLAÍ JIÈEWJÔQEÉE PÔNI LE BÚL"  \\
    Sub-branch & "EIN FOSTBALMTÉAM VÜYSPIYLYVT DIAS GEÛLBEN JÉAMÉ WÜÄACT DYN BLINT" & "QNE QPE DE FÚTBL JE JOUJEЕ À LA QPE JAJE PAXÙ LA BALE"  \\
    Language1 & "EIN FUSSBALLMATEAHMFT SPIELT DAS GELBE TEAM HAT DEN BALL" & "UNE ÉQUIPE DE FOOTBALL JOUE E LÉQUIPE JAUNE A LE BALLON" \\
    \hline[\heavyrulewidth]
    \textbf{Groundtruth} & "EIN FUSSBALLTEAM SPIELT DAS GELBE TEAM HAT DEN BALL" & "UNE ÉQUIPE DE FOOTBALL JOUE LÉQUIPE JAUNE A LE BALLON" \\
\end{tblr}
\caption{An example of the translation process (En-De and En-Fr) along the tree path of TET trained on Multi30K. 
The source English sentence is: ``\textit{A football team is playing, the yellow team has the ball}''. }
\label{tab:example0}
\end{table*}

\begin{table*}[ht]
\centering
\begin{tblr}{
  colspec = {Q[l] X[l] X[l] X[l]}, 
  width = \textwidth,
  hlines = {0.3pt},
  hline{1,Z} = {1.2pt},
  vlines = {0.3pt},
  vline{1,Z} = {0pt},
  rows = {t},
  row{1} = {font=\normalsize},
  row{2-Z} = {font=\small},
}
    \textbf{Tree Node} & \textbf{Dutch} & \textbf{French} & \textbf{Spanish} \\
    Common1 & "" & "" & "" \\
    Common2 & "" & "" & "" \\
    Branch & "" & "" & "" \\
    Sub-branch & "" & "" & "inclusive" \\
    Language1 & "Het" & "En" & "Al mismo tiempo inclusive el regiones no permitir ayudar esfuerzos esfuerzos resolución resolución conflictos conflicto" \\
    Language2 & "Tegelijkertijd zal het isolement van de afcheiden regio ' sningen voor opploslossing." & "Parallèlement, l ' isolement des régions dissidentistes ne contribuea pas les efforts de résolution des conflits." & "Al mismo tiempo, el aislamiento de las regiones no ayudará a los esfuerzos para resolver de conflictos." \\
    \hline[\heavyrulewidth]
    \textbf{Groundtruth} & "Tegelijkertijd draagt het isolement van de afgescheiden regio's niet bij aan de inspanningen voor het oplossen van het conflict." & "Dans le même temps, il convient de veiller à ne pas isoler les régions séparatistes, car cela ne ferait que saper les efforts déployés en vue de résoudre le conflit." & "Al mismo tiempo, el aislamiento de las regiones escindidas no contribuirá a los esfuerzos para la resolución del conflicto." \\
\end{tblr}
\caption{An example of translation process of TET trained on Europarl. The source English sentence is: ``\textit{At the same time, the isolation of the breakaway regions will not help efforts for conflict resolution.}''}
\label{tab:example1}
\end{table*}

\begin{table*}[ht]
\centering
\begin{tblr}{
  colspec = {Q[l] X[l] X[l] X[l]}, 
  width = \textwidth,
  hlines = {0.3pt},
  hline{1,Z} = {1.2pt},
  vlines = {0.3pt},
  vline{1,Z} = {0pt},
  rows = {t},
  row{1} = {font=\normalsize},
  row{2-Z} = {font=\small},
}
    \textbf{Tree Node} & \textbf{Dutch} & \textbf{French} & \textbf{Spanish} \\
    Common1 & "" & "" & "" \\
    Common2 & "" & "" & "" \\
    Branch & "" & "" & "" \\
    Sub-branch & "" & "" & "frecuentemente" \\
    Language1 & "De" & "Les et" & "inversiónversion inversión materia investigación agrícola y tecnologíacol verde inclusive son inversiónversion inversión futuro así pero crear nuevos puestos empleo" \\
    Language2 & "Investeringen in landbouw en groene technologie zijn investeringen in de toekomst en nieuwe banenen." & "Les investissements dans la recherche agricole et la technologies vertes sont des investissements dans l ' avenir et ils créeront de nouveaux emplois." & "Las inversiones en investigación agrícola y tecnologías ecológicas son inversiones en el futuro y crearán nuevos puestos empleo de trabajo." \\
    \hline[\heavyrulewidth]
    \textbf{Groundtruth} & "Investeringen in agrarisch onderzoek en groene technologie zijn investeringen in de toekomst en zullen nieuwe werkgelegenheid opleveren." & "Les investissements dans la recherche agricole et la technologie verte sont des investissements dans l'avenir et créeront de nouveaux emplois." & "Las inversiones en investigación agrícola y tecnologías ecológicas son inversiones de futuro y crearán nuevos puestos de trabajo." \\
\end{tblr}
\caption{An example of translation process of TET trained on Europarl. The source English sentence is: ``\textit{Investments in agricultural research and green technology are investments in the future and they will create new jobs.}''}
\label{tab:example2}
\end{table*}

\end{document}